\documentclass[11pt]{article}

\usepackage[preprint]{acl}

\usepackage{times}
\usepackage{latexsym}

\usepackage[T1]{fontenc}
\usepackage[utf8]{inputenc}

\usepackage{microtype}

\usepackage{inconsolata}

\usepackage{graphicx}

\title{Evaluating Token Probabilities as Confidence Estimators in Human Behavior Simulations: A Case Study in Luxembourgish Next Like Prediction}

\author{
  \textbf{Nils Schwager}$^1$, \textbf{Simon Münker}$^1$, \textbf{Alistair Plum}$^2$, \textbf{Achim Rettinger}$^1$ \\
  $^1$Trier University, Trier, Germany \\
  $^2$University of Luxembourg, Esch-sur-Alzette, Luxembourg \\
  \texttt{\{schwager, muenker, rettinger\}@uni-trier.de} \\
  \texttt{alistair.plum@uni.lu}
}
\begin{document}
\maketitle
\begin{abstract}
As large language models (LLMs) are increasingly used to simulate human behavior, their operational validity is typically assessed through accuracy alone, implicitly treating human behavior as deterministic. We argue that on noisy digital trace data, a model's performance depends not only on predicting the correct outcome but on its calibration. We present a case study testing whether the token probabilities of open-weight LLMs can serve as confidence estimates in a Next Like Prediction task: predicting how a user of the Luxembourgish news platform RTL.lu rates a comment, conditioned on their prior interactions. Evaluating Ministral 3 (3B, 8B), we find that, with in-context learning only, the tested models are overconfident and are outperformed on accuracy and calibration by a simple statistical baseline. Accuracy and calibration do not move together: scaling from 3B to 8B improves accuracy while degrading calibration, and longer behavioral histories raise accuracy while confidence declines. Task-specific supervised fine-tuning acts as an effective calibration intervention, reducing Expected Calibration Error from $0.2$ to $0.006$ while improving accuracy from $0.536$ to $0.758$. We conclude that confidence estimation should be evaluated alongside accuracy to improve validity when using LLM-based agents.
\end{abstract}

\section{Introduction}
As large language models (LLMs) are increasingly used to simulate human behavior, methodological research into the validity \cite{larooij2025validation, munker2026don} and alignment of generative agents \cite{chuang2024simulating, schwager2026towards} emphasizes that achieving operational validity requires alignment with authentic digital trace data. However, this type of data is inherently noisy due to the non-deterministic nature of human behavior and missing contextual modalities \cite{ouelletteHabitIntentionEveryday1998, zhangPuttingHumanBehavior2021}. This lack of information makes it difficult to achieve sufficient predictive performance, or even to determine confidence thresholds.

Contemporary research utilizing LLMs as generative agents mostly relies on evaluating alignment using discrete \cite{lu2026can, wang2026can} or continuous accuracy measures \cite{shen-etal-2025-valuecompass, schwager2026towards}, thereby treating human behavior as deterministic. We argue that in simulations (and LLM applications processing noisy and/or user-generated text in general), the LLM's utility depends not merely on its ability to predict the correct outcome, but on its capacity to accurately estimate its own predictive confidence \cite{farquharDetectingHallucinationsLarge2024, aichbergerRethinkingUncertaintyEstimation2025, nguyenProbabilitiesAreAll2026}. Because we will never reach absolute certainty on noisy data, we see calibration\footnote{We define \textit{calibration} as the degree of alignment between a model's confidence and its accuracy. Conversely, we use \textit{calibration intervention} to refer to the methodological procedure to align these probabilities with empirical observations - in our case, task-specific fine-tuning.} as a requirement for model alignment that is as important as accuracy itself \cite{azaria-mitchell-2023-internal, bakman-etal-2024-mars}.  

To address this gap, we present a case study investigating whether LLMs' token probabilities can natively serve as confidence metrics in human behavior simulation, and we evaluate task-specific fine-tuning as a calibration intervention. We utilize a \textbf{Next Like Prediction} task based on user comments and interactions from the Luxembourgish news platform RTL.lu\footnote{RTL  Lëtzebuerg: \url{https://www.rtl.lu/}}. Our task tests whether an \textbf{LLM can accurately predict a user's like or dislike of a given comment, based solely on previous (dis)likes.} The setup isolates calibration for two reasons: (A) it introduces severe linguistic complexity and contextual sparsity (Section~\ref{sec:methods:task_constraints}), and (B) it collapses human feedback into a strictly binary action space (like vs. dislike), which simplifies the measurement of accuracy and confidence relative to open-text alignment.

\subsection{Research Questions}
We investigate two research questions:
\begin{itemize}
    \setlength{\itemsep}{0pt}
    \setlength{\parskip}{0pt}
    \setlength{\parsep}{0pt}
    \item[$RQ_1$] How well do LLMs predict user likes/dislikes on Luxembourgish comments, and to what extent can we rely on their token probabilities as estimated confidence of these predictions?
    \item[$RQ_2$] How does task-specific fine-tuning impact both the discrete accuracy and the underlying calibration of the model's token probabilities?
\end{itemize}

\subsection{Contributions}
Our work makes the following contributions to the evaluation of LLM-based user modeling:

\paragraph{Confidence as an Evaluation Axis} 
We argue that alignment for generative agents should be assessed through calibration alongside accuracy, and we operationalize this by treating the normalized token probabilities of a binary decision as a native confidence estimate for a human behavior simulation task.

\paragraph{Accuracy and Calibration Divergence} 
Using a Next Like Prediction task on Luxembourgish digital trace data, we show that both tested off-the-shelf models are overconfident: they assign high probabilities to incorrect predictions and are outperformed on both accuracy and expected calibration error by a simple statistical baseline. Crucially, the two signals diverge: scaling from 3B to 8B improves accuracy while increasing calibration error, and lengthening the behavioral history improves accuracy while confidence declines. Especially in the first case, accuracy alone would misreport the model's probabilistic alignment.

\paragraph{Fine-Tuning as a Calibration Intervention} 
We show that task-specific supervised fine-tuning substantially reduces calibration error and widens the gap between confidence on correct and incorrect predictions, indicating that lightweight adaptation can bring a model's predictive probabilities into closer agreement with its observed accuracy.
\section{Background}
\label{sec:background}

\subsection{Luxembourgish NLP: Low Research but High Resource}
% intro 
Luxembourgish is the national language of the Grand Duchy of Luxembourg, and one of the four official languages alongside French, German, and German sign language. It is spoken natively by around 400,000 people \cite{Fehlen2023} in a society marked by extensive multilingualism and code-switching. These conditions, a small speaker base, no single fully codified written standard, and constant contact with two much larger (in terms of number of speakers) neighbouring languages, have long made Luxembourgish a difficult target for NLP \cite{purschke2025}. 

% start
A first strand of work on online discourse approached Luxembourgish through the lens of public attitudes towards multilingualism. Combining a Word2Vec based analysis of user comments from the RTL.lu news platform with crowdsourced questionnaire data collected through the Schn\"essen project, \citet{Purschke2020} connected sociolinguistic theories of language attitude with computational text analysis. A second strand similarly worked on the same online comments, but took an NLP approach in training various classifiers to moderate the comments \cite{ranasinghe-etal-2023-publish}. The work ultimately showed that multilingual models of the time worked well for this task, but lacked nuanced knowledge to effectively tackle problematic comments that should not be published.

% models
In terms of language models for Luxembourgish, \citet{Philippy2024} addressed the shortage of parallel data for Luxembourgish by compiling a small human generated cross-lingual parallel dataset, which was used to train LuxEmbedder, a sentence embedding model with cross-lingual capabilities. \citet{Lutgen2025} presented the first sequence to sequence normalisation models for Luxembourgish, built on the ByT5 and mT5 architectures and trained on word level data capturing real life spelling variation, in order to deal with the large amount of variation.

In addition to the growing model space for Luxembourgish \citep[see][]{Lothritz2022,Plum2025}, the recent availability of benchmarks \citep[see][]{Philippy2024c, Plum2026} highlights an important aspect of the current situation that Luxembourgish finds itself in with relation to NLP. Although \citet{joshi2020state} classified Luxembourgish as ``scraping-by'', and current research is ongoing but limited, the availability of data appears more as a ``high resource'' situation. Many sources of data are growing and becoming more readily available, including news articles and user comments from the main news broadcaster of Luxembourg, an expanding Wikipedia, transcripts of national assembly debates, online dictionary examples, to name but the largest collections. This difference in available data versus available methods highlights that processing and analysing the Luxembourgish language poses limiting difficulties, which are also apparent in the study at hand.

\subsection{Calibration in Generative Agents}
In machine learning, model calibration refers to the alignment of predictive probabilities (estimated confidence) with the observed accuracy \cite{geng2024survey, guo2017calibration}. 

\paragraph{Confidence in LLMs}
Applying traditional calibration methods directly to LLMs presents significant structural challenges. Unlike standard discriminative models, the output space for open-text generation grows exponentially with sequence length. While white-box methods attempt to estimate generation uncertainty using native logit scores, these indicators frequently suffer from surface form competition, where different valid textual representations compete for the same probability mass, skewing the actual confidence distribution \cite{holtzman2021surface}. Consequently, researchers have explored alternative white-box approaches, such as probing internal state activations \cite{kadavath2022language, li2023inference}. Black-box approaches require no access to the model's internal computational graph. Instead, these methods attempt to estimate uncertainty through consistency checks across multiple stochastic generations \cite{manakul2023selfcheckgpt}, evaluations via surrogate models \cite{shrivastava2023llamas}, or by directly prompting the model to verbalize its uncertainty, yielding a ``linguistic confidence'' score \cite{mielke2022reducing}.

\paragraph{Generative Agents}
% Shift towards LLMs
Recent methodological frameworks require the operational validity of generative agents to be established against authentic digital traces \cite{larooij2025validation, schwager2026towards}. We argue that for noisy, non-deterministic tasks such as simulating human behavior on digital trace data, quantifying calibration is a methodological requirement. Robust confidence estimation enables researchers to aggregate individual predictions and apply nuanced, probabilistic interpretations to computational social science simulations.

\paragraph{Research Gap}
This paper takes the probability a model attaches to a single predicted human action and checks it against whether that action occurred. Two lines of work approach that measurement and stop short. Calibration research assumes a resolvable ground truth, such as factual question answering or text labels with gold annotations \citep{gligoric-etal-2025-unconfident, tian-etal-2023-just}, so that a prediction can be scored against the correct answer. Behavior simulation gives up the probability instead of the ground truth. The evaluation of generative agents so far rests on believability or, more recently, on accuracy metrics \cite{larooij2025validation}. Neither asks whether a generative agent knows when it is wrong.

Human behavior offers no resolvable target: an observed action is one realization of a distribution over choices the individual could plausibly have made; like annotator disagreement on subjective labeling tasks, this variation is a property of the phenomenon rather than noise \citep{plank-2022-problem}. Because that uncertainty cannot be removed from the task, we begin by moving the evaluation of generative agents from discrete accuracy to probability: accuracy reports a determinate result for an outcome that was never determinate, whereas calibration asks whether its confidence matches its accuracy.

We therefore test calibration in generative agents on the binary prediction of user actions (likes or dislikes) over authentic, non-English, code-switched digital trace data; the binary format allows direct extraction of token probabilities as native confidence markers.
\section{Methods}
\subsection{Next Like Prediction Task}
\label{sec:methods:task}
The Next Like Prediction task centers on predicting how a specific user would respond to a given stimulus, in this case, evaluating whether they would \textit{like} (1) or \textit{dislike} (0) a comment. By comparing predicted interactions with authentic digital traces, we assess whether models can capture individual-level behavioral patterns rather than producing generic responses. This framing strictly follows the operational validity criterion: alignment should be measured against the actual individuals being simulated, not abstract notions of plausibility \cite{larooij2025validation, munker2026challenging}.

Following the methodology established by \citet{schwager2026towards}, we rely on implicit conditioning by providing the model with a behavioral history. Specifically, we format previous user interactions as preceding prompt-completion turns in the LLM's native chat structure. To standardize this conditioning and evaluate the impact of context scaling, the model receives a minimum of 3 historical interactions to predict the subsequent fourth interaction, with the context window expanding up to a maximum of $72$ interactions.  This effectively turns our prediction task into a classification task for a user's $(n+1)$-th interaction, based on the previous $n$ interactions. During evaluation, we strictly condition the model on the user's ground-truth behavioral history, explicitly omitting the model's own prior predictions to prevent compounding errors.

In addition to the behavioral history, the model is conditioned with the following system prompt: \textit{You are a social media user evaluating online posts. Keep your binary predictions (1 for like, 0 for dislike) consistent with your previous interaction history and the preferences you have expressed earlier}.

\subsection{Dataset}
The data for this task is provided by RTL.lu, the main news broadcaster of Luxembourg and the only one entirely in Luxembourgish. On the \url{rtl.lu} website, registered users can post comments below journalistic news articles. Users can react to the comments of other users with a \textit{like} or a \textit{dislike}. We are provided this data, which allows us to reconstruct which user liked or disliked which comments.\footnote{At the time of writing this article, we are unable to share the data due to privacy agreements. We hope that in the future, a privacy protected version may be made publicly available.} The oldest comment in the dataset is from March 2020 and the newest from May 2024.

\paragraph{Pre-Processing}
First, we reorganize reactions by user to enable user-level modeling. We model the users strictly as voters; the stimuli are comments made by other users. To prevent overfitting to power users, we remove the top 20\% of users by interaction count. This leaves users with at most 72 stimulus–response pairs. Users with fewer than four interactions are excluded. 

\paragraph{Splits \& Size}
We partition data at the user level so that all stimulus–response pairs from a single user appear exclusively in training or testing. This prevents cross-user leakage and enables the evaluation of cross-user generalization. Our resulting training set comprises $3,247$ users and $57,185$ training cases (excluding the three required seed conditioning cases), as well as $783$ test users and $13,759$ test cases. 

\paragraph{Descriptive Statistics}
The filtered dataset exhibits a baseline class distribution of 61\% likes and 39\% dislikes, establishing the base rate for the classification task. Within the defined bounds, users have an average of $20.5$ interactions (median = $13$). Furthermore, the user comments have an average length of $49$ words. These comments are written in Luxembourgish (75\%), English (15\%), French (4\%), and German (3\%), reflecting the multilingual user base of the platform.

\subsection{Task and Data Constraints}
\label{sec:methods:task_constraints}

To contextualize the task difficulty, we note that the accuracy for this prediction task is inherently bounded by ambiguity in the digital trace data, stemming from three primary sources:

\paragraph{Linguistic Complexity}
The dataset contains mainly Luxembourgish, which also appears in mixed and unmixed combinations with German, French, and English. Furthermore, colloquial Luxembourgish exhibits significant variation and code-switching. This linguistic diversity creates a challenging setting, which also tests the models' different language capabilities.

\paragraph{Contextual Sparsity}
The prediction task relies exclusively on a sparse historical context window ranging from $3$ to $72$ previous interactions. Because the parent journalistic articles and broader external contexts are deliberately omitted from the input, the model must infer the initial topic solely from the target comment text. The model must subsequently derive a generalized user preference from this construct to execute the prediction.

\paragraph{Behavioral Non-Determinism}
Our data lacks information on latent variables that heavily influence behavior, such as offline events. Further, we assume human interaction is fundamentally non-deterministic; a user might upvote a comment on one day but ignore a similar one on another. Because these latent factors are collapsed into a binary outcome, perfect prediction is not expected \cite{hofmanPrediction}. Consequently, this task focuses on evaluating the model's ability to appropriately reflect its confidence given the noise.

\subsection{Metrics}
We use standard accuracy as a discrete measure and treat the token probabilities for the tokens 0 and 1 as the model's confidence. We acknowledge that interpreting token probabilities as confidence is bounded by factors such as first-token deviations \cite{wang2024my} and surface form competition \cite{holtzman2021surface}. However, this limitation is mitigated in our setup because the model strictly predicts discrete states (0 or 1) and is explicitly conditioned to enforce format compliance (Section~\ref{sec:methods:task}). We apply a standard softmax function ($t=1$) over the full vocabulary and extract the probabilities for our target tokens (0 and 1). In a second step we normalize the probability mass of 0 and 1 to ensure they form a valid probability distribution over the binary target space.

\paragraph{Expected Calibration Error (ECE)} 
To quantify the alignment between the model's confidence and empirical reality, we calculate the Top-1 ECE \cite{guo2017calibration}. We partition the normalized predictions into $M=10$ equally spaced confidence bins. ECE measures the weighted average of the absolute difference between the empirical accuracy and the mean predicted confidence within each bin, thereby penalizing models that are systematically over- or underconfident.

\paragraph{Brier Score} 
The Brier score evaluates the mean squared difference between the model's predicted probabilities and the actual binary outcomes \cite{VERIFICATIONOFFORECASTSEXPRESSEDINTERMSOFPROBABILITY}. Unlike discrete accuracy, the Brier score penalizes extreme overconfidence on incorrect predictions.

\paragraph{ROC-AUC} 
To measure the model's underlying discriminative power independent of a rigid $0.5$ decision threshold, we report the Area Under the Receiver Operating Characteristic Curve (ROC-AUC). This evaluates the probability that the model assigns a higher token probability to a randomly chosen positive interaction (like) than to a randomly chosen negative one (dislike).

\paragraph{Conditional Confidence} 
To diagnose the calibration deficit, we disaggregate the model's certainty into two complementary metrics: Confidence on Correct Predictions ($\text{Conf}_{\text{Correct}}$) and Confidence on Incorrect Predictions ($\text{Conf}_{\text{Incorrect}}$). A model whose confidence is informative will exhibit a large gap between these values: high certainty when accurate, and appropriate uncertainty (probabilities approaching $0.5$) when predicting incorrectly.

\subsection{Experimental Setup}

\paragraph{Model \& Fine-Tuning}
We utilize the open-weight Ministral 3 model family (\texttt{Ministral-3-3B-Instruct-2512-BF16} and \texttt{Ministral-3-8B-Instruct-2512-BF16} \cite{liu2026ministral}) because its robust pre-training on diverse European languages establishes a strong linguistic baseline for Luxembourgish text, while remaining parameter-efficient. We apply Supervised Fine-Tuning (SFT) to the models on the task described in Section~\ref{sec:methods:task}. To ensure comparability across models, we use identical hyperparameters: one epoch, a maximum sequence length of $4,500$ tokens, and training on complete input sequences (system prompt, user prompts, and model completions). We use the paged AdamW optimizer with 8-bit quantization \cite{dettmers20218} to enable training on a single NVIDIA L40S GPU (48GB VRAM). All remaining hyperparameters follow the TRL defaults \cite{vonwerra2022trl}.

\paragraph{Statistical Baseline} 
To establish a non-semantic lower bound, we implement a naive statistical baseline. For each prediction step, the probability of a positive interaction is computed strictly as the distribution of the user's prior interactions (total past likes divided by total historical interactions). 

\paragraph{Proprietary Baseline} 
To establish a baseline of proprietary model capabilities, we evaluate \texttt{Gemini 3.5 Flash-Lite} (with reasoning disabled), a recent high-throughput model optimized for latency-sensitive classification tasks. As with most SOTA proprietary models, the internal logits remain inaccessible via the standard API, hence we only report its discrete predictive accuracy.
\section{Results}
\begin{table*}[t]
\centering
\small
\setlength{\tabcolsep}{4.5pt}
\begin{tabular}{@{}lcccccc@{}}
\toprule
\textbf{Model} & \textbf{Accuracy ($\uparrow$)} & \textbf{ROC-AUC ($\uparrow$)} & \textbf{Brier Score ($\downarrow$)} & \textbf{ECE ($\downarrow$)} & \textbf{Conf Cor. ($\uparrow$)} & \textbf{Conf Incor. ($\downarrow$)} \\
\midrule
Statistical Baseline & 0.5900 & 0.5488 & 0.2476 & 0.0642 & 0.6581 & \textbf{0.6346} \\
Ministral 3 3B & 0.4906 & 0.6128 & 0.2813 & 0.1650 & 0.6632 & 0.6482 \\
Ministral 3 3B (SFT) & 0.7089 & 0.7600 & 0.1880 & 0.0087 & 0.7303 & 0.6434 \\
Ministral 3 8B & 0.5355 & 0.5872 & 0.2937 & 0.2004 & 0.7494 & 0.7202 \\
Ministral 3 8B (SFT) & \textbf{0.7578} & \textbf{0.8253} & \textbf{0.1618} & \textbf{0.0056} & \textbf{0.7828} & 0.6616 \\
Gemini 3.5 Flash Lite & 0.6868 & - & - & - & - & - \\
\bottomrule
\end{tabular}
\caption{\textbf{Comparative Model Performance and Calibration.} Results detail discrete accuracy and probabilistic calibration across all evaluated configurations. Best values per metric are \textbf{bolded}.}
\label{tab:calibration_metrics}
\end{table*}
\label{sec:results}
\subsection{Model Calibration ($RQ_1$)}
\begin{figure}[!b]
    \centering
    \includegraphics[width=\linewidth]{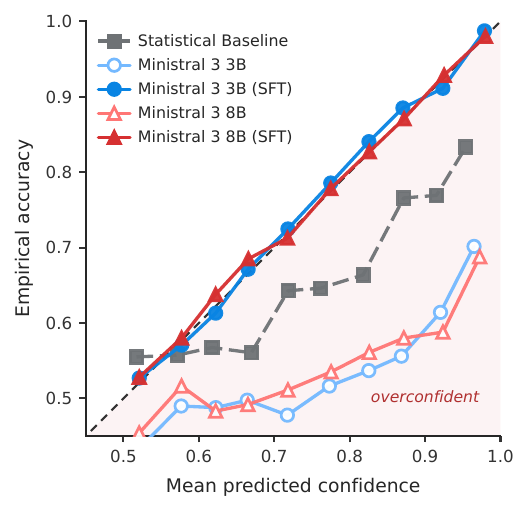}
    \caption{Reliability diagram comparing model calibration on the Next Like Prediction task.}
    \label{fig:calibration_diagram}
\end{figure}
To address $RQ_1$, we first evaluate the off-the-shelf models to observe how well they estimate their own certainty on noisy digital trace data. The results, detailed in Table~\ref{tab:calibration_metrics} and visualized in the reliability diagram (Figure~\ref{fig:calibration_diagram}), show that these LLMs are overconfident in the tested setting.
When evaluated on authentic user interactions, the 8B off-the-shelf model achieves a discrete accuracy of $0.5355$, performing only slightly above random chance. Notably, this underperforms our simple statistical baseline, which achieves an accuracy of $0.5900$ by strictly predicting the user's historical marginal distribution. Despite this low predictive accuracy, the 8B model exhibits a high ECE of $0.2004$ and a Brier score of $0.2937$, a penalty worse than simply predicting a constant $0.5$ probability for every interaction. 

This overconfidence becomes clear when looking at the conditional confidence metrics. When the 8B model makes an incorrect prediction, it still assigns a mean probability of $0.7202$ ($\text{Conf}_{\text{Incorrect}}$) to its choice. This is nearly as high as its confidence when it is correct ($\text{Conf}_{\text{Correct}} = 0.7494$). In contrast, the purely statistical baseline naturally avoids this overconfidence, maintaining a much lower $\text{Conf}_{\text{Incorrect}}$ of $0.6346$. The 3B model behaves similarly to the 8B, demonstrating a $\text{Conf}_{\text{Incorrect}}$ of $0.6482$ against an overall accuracy of only $0.4906$.

\subsection{Fine-Tuning and Calibration Improvement ($RQ_2$)}
To answer $RQ_2$, we examine the impact of task-specific SFT on the models' probability outputs. We find that fine-tuning improves the models' ability to appropriately express uncertainty while simultaneously increasing their discrete performance.

As shown in Table~\ref{tab:calibration_metrics}, fine-tuning the 8B model increases predictive accuracy to 0.7578 and raises ROC-AUC to $0.8253$. This gain over the statistical baseline suggests that the model learns semantic behavioral patterns rather than defaulting to a user's historical average. The same intervention aligns the model's predicted probabilities more closely with its observed accuracy: fine-tuning reduces the ECE to 0.0056 and lowers the Brier score to $0.1618$.

This improvement is also visible in the conditional confidence gap. For the fine-tuned 8B model, mean confidence on correct predictions rises to $0.7828$, while confidence on incorrect predictions falls to $0.6616$. This widening gap is consistent with a model that lowers its probability estimates when its predictions are more likely to be wrong.

Finally, we observe that achieving good calibration does not strictly require larger models. The 3B (SFT) model achieves calibration nearly on par with the 8B (SFT) model (ECE $0.0087$ vs. $0.0056$), indicating that strong calibration is attainable at smaller scale, although the 8B (SFT) model remains best on both accuracy and calibration. Overall, task-specific fine-tuning proves to be effective not only for predicting user interactions but also for ensuring the model's confidence scores are sufficiently calibrated for downstream analysis.

When evaluated on the same dataset, the proprietary \texttt{Gemini 3.5 Flash-Lite} model achieves a discrete accuracy of $0.6868$. While this outperforms the statistical baseline and off-the-shelf models, it falls short of our fine-tuned 8B model. 

\subsection{Contextual Scaling Analysis}

\begin{table}[!htbp]

\centering

\footnotesize
\setlength{\tabcolsep}{4pt}
\begin{tabular}{@{}lcccc@{}}

\toprule

\textbf{Model} & \multicolumn{2}{c}{\textbf{History $\to$ Acc}} & \multicolumn{2}{c}{\textbf{History $\to$ Conf}} \\
\cmidrule(lr){2-3} \cmidrule(l){4-5}
 & $\Delta_{100}$ & $r$ & $\Delta_{100}$ & $r$ \\
\midrule
Statistical Baseline & 0.195 & $0.49^{***}$ & -0.017 & $-0.20$ \\
Ministral 3 3B & 0.100 & $0.23$ & -0.068 & $-0.71^{***}$ \\
Ministral 3 3B (SFT) & 0.095 & $0.34^{**}$ & 0.050 & $0.49^{***}$ \\
Ministral 3 8B & 0.247 & $0.50^{***}$ & -0.087 & $-0.64^{***}$ \\
Ministral 3 8B (SFT) & 0.086 & $0.33^{**}$ & 0.045 & $0.43^{***}$ \\
Gemini 3.5 Flash Lite & 0.143 & $0.43^{***}$ & - & - \\
\bottomrule

\end{tabular}

\caption{\textbf{Contextual Scaling.} $\Delta_{100}$ represents the linear change in accuracy and confidence per 100 interactions, rescaled from a slope fitted over the observed range (at most 72 interactions) and $r$ denotes the Pearson correlation coefficient. Significance levels: $^{*}p<0.05$, $^{**}p<0.01$, $^{***}p<0.001$.}
\label{tab:context_scaling}

\end{table}

As a supplementary analysis, we examine accuracy and confidence as a function of behavioral history length. 
The results, detailed in Table~\ref{tab:context_scaling}, reveal a divergence between accuracy and confidence in the off-the-shelf models. While providing extended behavioral history significantly improves the accuracy of the 8B model ($r=0.50$), its internal confidence exhibits a strong inverse trajectory ($r=-0.64$). As the context window expands, the model becomes more accurate but less confident.

Conversely, Supervised Fine-Tuning (SFT) realigns these metrics. For the fine-tuned 8B model, both accuracy ($r=0.33$) and confidence ($r=0.43$) scale positively and significantly with history length. This indicates that task-specific adaptation trains the model to utilize extended digital traces not just for discrete prediction, but to increase its confidence. 

Finally, while the purely statistical baseline and the proprietary \texttt{Gemini 3.5 Flash-Lite} model both achieve significant accuracy gains with longer historical contexts ($r=0.49$ and $r=0.43$, respectively), their usable confidence signal is constrained. The baseline's confidence shows no significant association with history length ($r=-0.20$).

For the off-the-shelf models, this supports our central claim that accuracy alone is an insufficient evaluation criterion, since accuracy and the confidence signal move in opposite directions.
\section{Discussion}

\subsection{Interpreting the Calibration Deficit}
The results from Section~\ref{sec:results} highlight a divergence between the models' discrete predictive capacity and their calibration. First, we observe that the statistical baseline natively avoids overconfidence because it anchors its predictions strictly to the distribution of historical interactions. This results in the lowest mean probability assigned to incorrect predictions ($0.6346$). In contrast, the off-the-shelf models exhibit a calibration deficit, assigning high probability mass to incorrect outcomes, which inflates their Brier scores.

Second, our findings indicate that accuracy and calibration do not inherently scale together. Increasing the parameter count from the 3B to the 8B model improves accuracy (from $0.4906$ to $0.5355$) but increases the ECE from $0.1650$ to $0.2004$. This decoupling demonstrates that parameter scaling alone does not resolve calibration deficits. 

Third, SFT functions as an effective calibration intervention. Task-specific fine-tuning closes the gap between mean confidence and accuracy, and widens the gap between $\text{Conf}_{\text{Correct}}$ and $\text{Conf}_{\text{Incorrect}}$. Following SFT, the 8B model increases its confidence on correct predictions to $0.7828$, while reducing it on incorrect predictions to $0.6616$.

\subsection{The Insufficiency of Inference-Only Paradigms}
Common NLP downstream tasks can currently be solved by relying on in-context learning coupled with few-shot examples \cite{brown2020language}. This narrative of universal adaptability without requiring data-driven weight updates is frequently applied to user-behavior generation and prediction tasks \cite{larooij2025validation}. However, authentic digital trace data, especially non-English and code-switched text, deviates heavily from original pre-training distributions. 

Our results demonstrate a discrepancy between the performance of an in-context, prompt-based approach and data-driven fine-tuning. The empirical finding that a naive statistical heuristic achieves superior calibration and accuracy compared to off-the-shelf models provides evidence against the assumption that generalized inference is universally sufficient for human simulation. 

If a simple mathematical baseline is cheaper, more transparent, and achieves a safer confidence distribution than unaligned open-weight models, utilizing off-the-shelf LLMs via standard inference is inadequate for rigorous modeling \cite{shen2025bidirectional}. Therefore, our results suggest that training parameter-efficient models and adapting them specifically for the individual tasks is a methodological requirement.

\subsection{Accessibility needed?}
Measuring probabilistic alignment, which we deem relevant for evaluation of behavior simulations, requires unrestricted access to a model's internal probability mass. For instance, while the proprietary \texttt{Gemini 3.5 Flash-Lite} achieves a competitive discrete accuracy of $0.6868$ on our dataset, its restricted API obscures the underlying logit scores. Consequently, calibration metrics like ECE or Brier scores cannot be computed. 

This limitation reduces black-box evaluation to deterministic point-estimates, without associated uncertainty. For rigorous downstream evaluation, accessing the raw computational graph is a methodological prerequisite. Future work must bridge this gap by applying mechanistic interpretability methods \cite{dunefsky2024transcoders}, such as circuit analysis or sparse autoencoders, to open-weight architectures to formally dissect how generative agents construct confidence.

\subsection{The Case for Confidence in Generative Agents}
We make the case that when benchmarking the ecological and operational validity of generative agents, researchers must explicitly test for confidence alongside accuracy. However, we acknowledge that current methods for extracting confidence in open-text generation are insufficient and under-researched, particularly within the intersection of Computational Social Science and LLMs. 

Despite these current methodological limits, shifting from discrete deterministic accuracy to probabilistic evaluation is necessary. Establishing standardized calibration approaches would allow researchers to quantify the robustness of simulations by aggregating individual confidence estimates. Ultimately, recognizing and evaluating the inherent uncertainty in human digital trace data will lead to more sound and transparent computational social science applications.
\section{Conclusion}
% I need to rework this and the following sections.
\paragraph{$RQ_1$: LLMs on the Task}
Our empirical evaluation demonstrates that off-the-shelf models show lower performance than naive statistical estimations based on the marginal like/dislike distributions, a baseline neither off-the-shelf model beats on accuracy or on calibration. This discrete inaccuracy is accompanied by a confidence signal that does not track it: across both model scale and context length, accuracy and confidence signal move in opposite directions, so higher accuracy alone does not result in better probabilistic alignment.

\paragraph{$RQ_2$: Impact of Fine-Tuning}
The impact of task-specific fine-tuning is so significant that data-driven alignment appears necessary for the noisy task at hand. It marks the difference between underperforming the statistical baseline via in-context learning, and outperforming the baseline significantly after training the model on the task. Crucially, it also shows the overarching benefit of achieving closer agreement between measured confidence and observed accuracy.

\subsection{Future Work}
\paragraph{Expanding to Open Text}
Our current methodology strictly bounds the action space to a binary classification task to extract and isolate discrete token probabilities. Expanding this framework to generative, open-text simulations requires advancing beyond single-token estimations to accurately capture the behavioral diversity in authentic digital traces.

\paragraph{Advanced Approaches for Modeling Uncertainty}
To achieve this, future research must evaluate the calibration of generative agents using more complex paradigms for modeling uncertainty. To circumvent the limitations of discrete token probabilities, internal state-based methods present a promising research direction.

\paragraph{Multi-Turn Calibration}
Currently, our evaluation is constrained to single-turn predictions conditioned exclusively on empirical ground-truth histories. To facilitate complex computational social simulations, models must be capable of multi-turn interactions, autoregressively conditioning on their own synthetic outputs. Future research must evaluate calibration in these dynamic environments, specifically investigating how to quantify and mitigate the compounding uncertainty that occurs during sequential generation.

\paragraph{Testing on Out-of-Distribution (OOD) Data}
Addressing these structural challenges paves the way for robust, ecologically valid generative agents. Ultimately, the community must benchmark probabilistic alignment on out-of-distribution data, as the generalizability of current models across diverse behavioral patterns remains uncertain. Because a truly operational simulation must reliably generate insights across counterfactual scenarios, achieving robust confidence estimates on OOD data during multi-turn interactions remains the central open challenge in human behavior simulation. 

\section*{Limitations}

\paragraph{Generalizability Across Models and Domains} As a case study, our findings are bounded to a single open-weight model family (Ministral 3) evaluated on a single dataset. While this focus enables a controlled analysis of scale and fine-tuning within one architecture, the overconfidence we observe is established for this specific setting rather than for open-weight LLMs in general. Consequently, the calibration deficit and its correction under fine-tuning do not automatically translate to other model families or other forms of noisy behavioral data, which we leave to future work.

\paragraph{Constraint to Discrete Action Spaces}
A primary limitation of our current methodological framework is its strict reliance on a bounded, binary action space. While scaling this predictive paradigm to multi-class discrete states is mathematically straightforward, quantifying confidence in open-ended text generation introduces complexity. In unconstrained generative environments, foundational metrics such as perplexity serve as inadequate proxies for uncertainty. Consequently, our findings regarding token-level calibration on single-token outputs do not automatically translate to the confidence estimation of free-text behavioral simulacra.

\paragraph{Single-Turn Evaluation Scope}
Furthermore, our current experimental design is constrained to single-turn inference, evaluating predictions conditioned exclusively on empirical, ground-truth historical interactions. This isolates baseline calibration but leaves multi-turn performance unaddressed. We do not evaluate the model's capacity to maintain probabilistic alignment when autoregressively conditioning on its own synthetic prior outputs. The compounding nature of uncertainty over sequential generations remains a critical unknown; it is not yet established how to calibrate a multi-turn model once its contextual history inevitably diverges from authentic digital trace data.
\section*{Ethical Considerations}

\paragraph{Behavioral Prediction as a Dual-Use Capability}
Our task, by design, asks an LLM to infer a specific, identifiable person's future endorsement from a fragment of their prior digital behavior. Framed as a calibration study, this is a methodological probe; framed outside that context, it is a template for individual-level preference profiling. Our central finding, that fine-tuning not only raises accuracy but also sharpens confidence, is what would make such a system attractive for downstream deployment (e.g., comment ranking, engagement optimization, or targeted content curation). A well-calibrated predictor of what a person will approve of is a more trustworthy-looking, and therefore more readily deployable, tool for behavioral targeting than an overconfident one. We see the same calibration intervention we advocate for methodologically as one that lowers the barrier to real-world use of this capability.
 
\paragraph{Miscalibrated Confidence as a Deployment Risk}
Beyond privacy, our results affect research and deployment practices: off-the-shelf models are not only inaccurate but also assign high confidence to their own errors (Section~\ref{sec:results}). Suppose such models were adopted for behavioral simulation or moderation-adjacent decisions without the calibration intervention step we test here, their incorrect but confident outputs risk being trusted as reliable when they are not. Conversely, evaluating LLMs on this kind of task using in-context learning alone, without a transparent statistical baseline or a fine-tuned comparison point, risks the opposite error: concluding that LLMs are broadly unsuited to modeling noisy human behavior when the deficit lies in the inference-only paradigm rather than in the underlying model. Reporting calibration alongside accuracy, and reporting both off-the-shelf and fine-tuned results against a naive baseline as we do, helps guard against both of these premature conclusions.
 
\paragraph{Positioning within Low-Resource NLP}
Luxembourgish remains under-resourced for NLP despite growing data availability (Section~\ref{sec:background}). Contributing an evaluation of confidence estimation on authentic Luxembourgish digital trace data supports this community's research infrastructure and, in principle, downstream applications such as improved comment moderation tooling for the platform that provided the data. We consider this a genuine benefit that should be weighed against the risks above, but not one that offsets them by default: the same methodology, applied to a larger or less consenting population, would carry the profiling risks described above with fewer of the mitigating factors (small-scale, agreement-bound, non-public data) present in our setup. 
\section*{Acknowledgements}
This work was conducted within the Ecological Validation of Artificial Simulacra (EVAS) Research Cooperative (\url{https://evas-research.github.io/}). This work is supported by EMO-SCI (project number 569156112) funded by Deutsche Forschungsgemeinschaft (DFG, German Research Foundation). We thank RTL for providing us with the data for research purposes.
\bibliography{custom}

\appendix
\section{Code and Model Availability}
The technical pipeline and source code are publicly available in the following repository: \url{https://github.com/nsschw/accuracy-is-not-enough}. To mitigate potential misuse while ensuring reproducibility, fine-tuned models are restricted to scientific use and shared only upon request. This policy aligns our open science commitment with responsible research practices.

\end{document}